\documentclass[11pt,a4paper]{article}
\usepackage[hyperref]{emnlp2020}

\usepackage{url}

\usepackage{tabu} 
\usepackage{bbm}
\usepackage{bm}
\usepackage{transparent}
\usepackage{CJKutf8}
\usepackage{latexsym}
\usepackage{xcolor}

\usepackage{microtype}
\usepackage{balance}

\usepackage{tikz}
\usepackage{tikz-qtree}

\usepackage{graphicx}			
\usepackage{times}

\usepackage{mathpartir}
\usepackage{pifont}

\usepackage{proof}
\usepackage{xspace}
\usepackage{multirow}
\usepackage{forest}
\usepackage{subcaption}

\usepackage{dsfont}
\usepackage{amsmath}
\usepackage{amssymb}
\usepackage{amsthm}
\usepackage{algorithm}
\newcommand{\namecite}[1]{\newcite{#1}}

\newcommand{\notes}[1]{}

\theoremstyle{definition}

\theoremstyle{plain}

\newcommand{\vecz}{\ensuremath{\mathbf{z}}}
\newcommand{\vech}{\ensuremath{\bm{h}}\xspace}

\newcommand\dbar[1]{\overline{\overline{#1}}}

\newcommand{\ith}[1]{\ensuremath{i^{{th}}}}

\newcount\permx
\newcount\permy
\def\permdot#1#2{
\permx=#1 \advance\permx by-1
\permy=#2 \advance\permy by-1
\psframe[fillcolor=black, fillstyle=solid]
(\permx,\permy)(#1, #2)
}

\newcommand{\argmax}{\operatornamewithlimits{\mathbf{argmax}}}

\newcommand{\tb}{\ensuremath{{\mathit{TB}}}\xspace}

\newcommand{\boxnum}[1]{{\setlength{\fboxsep}{1pt}\raisebox{1pt}{\hspace{1pt}\fbox{\tiny #1}\hspace{1pt}}}}
\newcommand{\ind}[1]{\ensuremath{_{\kern-0.5pt\boxnum{#1}}}}

\newcommand{\xbar}{\ensuremath{\overline{x}}\xspace}
\newcommand{\wbar}{\ensuremath{\overline{w}}\xspace}
\newcommand{\xbarbar}{\ensuremath{\dbar{x}}\xspace}
\newcommand{\wbarbar}{\ensuremath{\dbar{w}}\xspace}

\newcommand{\vecy}{\ensuremath{{\bm{y}}}\xspace}
\newcommand{\vecybar}{\ensuremath{\overline{\vecy}}\xspace}

\newcommand{\vecw}{\ensuremath{\bm{w}}\xspace}
\newcommand{\vecc}{\ensuremath{\mathbf{c}}\xspace}

\def\namecite{\newcite}

\newcommand{\smallnt}[1]{\ensuremath{_{\mbox{\tiny PP}}}\xspace}

\newcommand{\pseudocode}{Algorithm}
\floatname{algorithm}{\pseudocode}

\iffalse

\else

\fi

\newcommand{\eos}{\mbox{\scriptsize \texttt{<eos>}}\xspace}

\definecolor{chocolate}{rgb}{0.28, 0.02, 0.03}
\definecolor{PaleGreen}{rgb}{0.33, 0.545,0.33}
\definecolor{colorC0}{RGB}{51,113, 169}
\definecolor{colorC1}{RGB}{243,130,37}
\usepackage{array}
\usepackage{diagbox}

\aclfinalcopy

\title{Incremental Text-to-Speech Synthesis with Prefix-to-Prefix Framework
\thanks{\; 
  \ifaclfinal
  M.~M.~and B.~Z.~contributed equally;
  M.~M.~co-directed the project (with L.~H.), and was responsible for the majority of ideas and implementations;
  B.~Z.~improved the speech quality significantly and implemented the ideas on different TTS systems.
  See our generated audio samples and demos at 
 {\url{https://inctts.github.io/}}. 
  $^\mathsection$Currently address: Kwai Inc., Seattle, WA.
  $^\mathparagraph$Current address: Amazon, San Francisco, CA.
  \fi
 } 
       \ifaclfinal
         \else
       \fi
       }
\author{Mingbo Ma$^{\dagger *}$   \,
Baigong Zheng$^{\dagger \mathsection *}$\,
Kaibo Liu$^{\dagger}$ \,
Renjie Zheng$^{\dagger}$\,
\\
  {\bf
Hairong Liu$^\dagger$ \,
Kainan Peng$^{\dagger \mathparagraph}$ \,
Kenneth Church$^\dagger$ \,
Liang Huang$^{\dagger,\ddagger}$ \,
}
\\[0.1cm]
  $^{\dagger}$Baidu Research, Sunnyvale, CA, USA \;\; $^\ddagger$Oregon State University, Corvallis, OR, USA
\\
  {\tt \small \{mingboma, lianghuang\}@baidu.com }
}
\date{}

\begin{document}
\maketitle
\begin{abstract}
  Text-to-speech synthesis (TTS) has witnessed rapid progress in recent years,
  where neural methods became capable of producing audios with high naturalness.
  However, these efforts still suffer from two types of latencies:
  (a) the {\em computational latency} (synthesizing time), which grows linearly with the sentence length,
  and
  (b) the {\em input latency} in scenarios where the input text is incrementally available (such as in simultaneous translation,
  dialog generation, and assistive technologies).
  To reduce these latencies, we propose a neural incremental TTS approach 
  using the prefix-to-prefix framework from simultaneous translation. 
  We synthesize speech in an online fashion, playing a segment of audio while generating the next,
  resulting in an $O(1)$ rather than $O(n)$ latency.
Experiments on English and Chinese TTS show that our approach achieves 
similar speech naturalness compared to full sentence TTS,
but only with 
a constant (1--2 words) latency.

\end{abstract}

\section{Introduction}
Text-to-speech synthesis (TTS) generates speech from text, and is an important task with wide applications
in dialog systems, speech translation, natural language user interface, assistive technologies, etc.
Recently, it has benefited greatly from deep learning,
with neural TTS systems becoming capable of generating audios with high naturalness \cite{oord+:2016,shen+:2018}.

\begin{figure}[t]
\centering
\includegraphics[width=.99\linewidth,height=4cm]{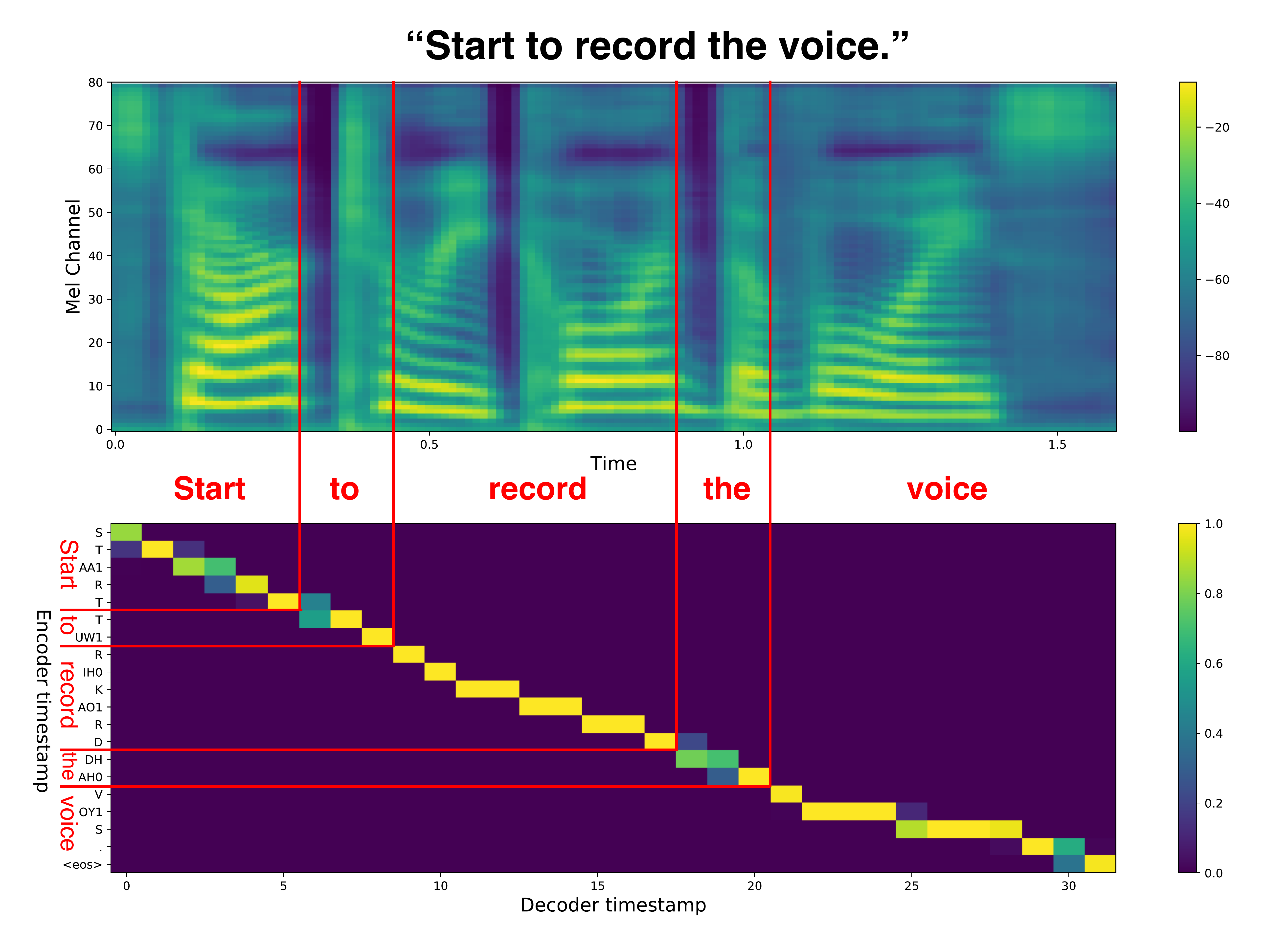}
\vspace{-0.7cm}
\caption{Monotonic spectrogram-to-text attention.\vspace{-.3cm}}
\label{fig:monoattn}
\vspace{-0.2cm}
\end{figure}

State-of-the-art neural TTS systems generally 
consist of two stages: 
the {\em text-to-spectrogram} stage which generates
an intermediate acoustic representation (linear- or mel-spectrogram) from the text,
and the {\em spectrogram-to-wave} stage (vocoder)
which converts the aforementioned acoustic representation into actual wave signals.
In both stages, there are sequential approaches based on the seq-to-seq framework, as well as more recent parallel methods. 
The first stage, being relatively fast,
is usually sequential \cite{wang+:2017,shen+:2018,li+:2019}
with a few exceptions \cite{ren+:2019,peng+:2019},
while the second stage, being much slower,
is more commonly parallel \cite{oord+:2018,prenger+:19}.

\begin{figure*}[t]
\centering
\includegraphics[width=.99\linewidth]{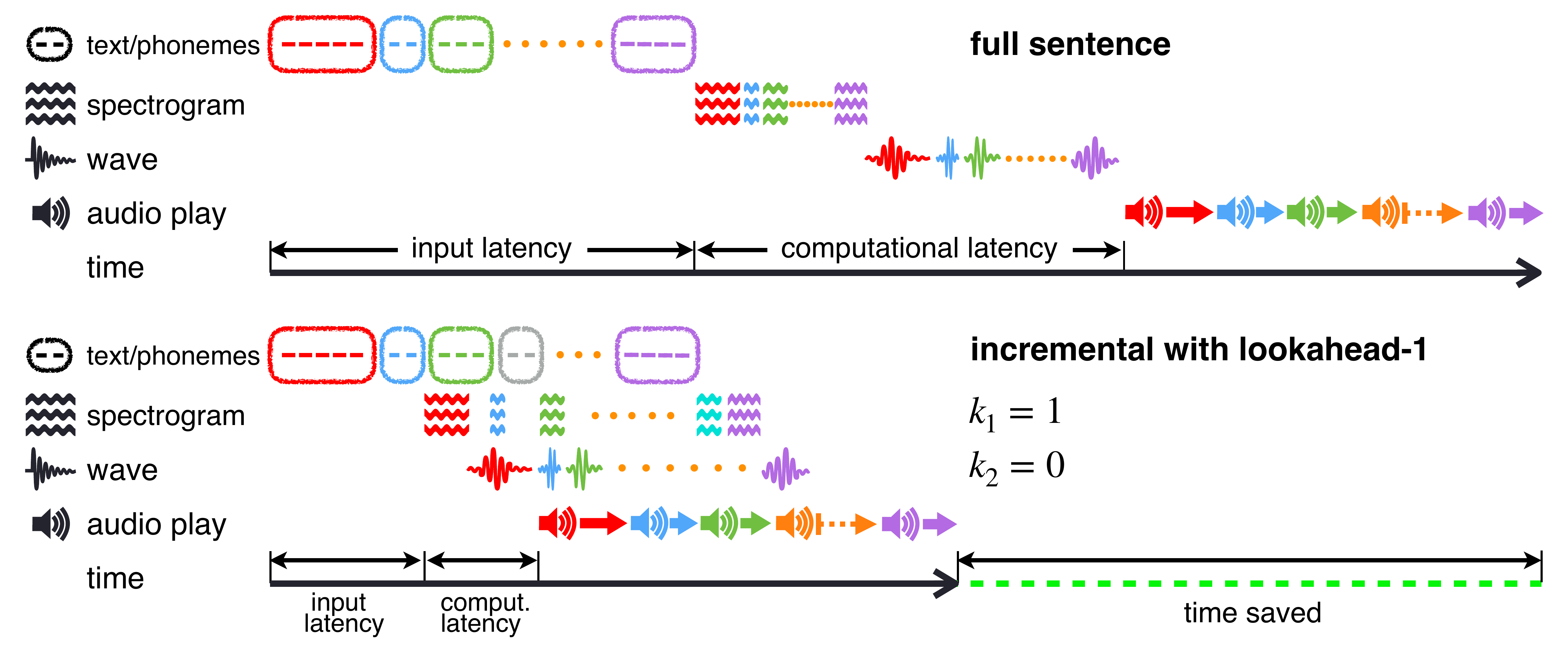}
\caption{Full-sentence TTS vs. our proposed incremental TTS with prefix-to-prefix framework (with $k_1=1$ and $k_2=0$ in Eq.~\ref{eq:waitk}). Our increnemtnal TTS has much lower latency than full-sentence TTS.
Our idea can be summarized by a Unix pipeline:
{\tt\small cat text | text2phone | phone2spec | spec2wave | play} (see also Fig.~\ref{fig:pipeline}),
where different modules can be processed parallelly.
}
\label{fig:p2p}
\vspace{-0.4cm}
\end{figure*}

Despite these successes,
standard full-sentence neural TTS systems still suffer from
two types of latencies:
(a) the {\em computational latency} (synthesizing time),
which still grows linearly with the sentence length even using parallel inference
(esp.~in the second stage),
and (b) the {\em input latency}
in scenarios where the input text is incrementally generated or revealed,
such as in simultaneous translation \cite{bangalore+:2012,ma+:2019},
dialog generation \cite{skantze+:2010,buschmeier+:2012}, and assistive technologies \cite{elliott:2003}.
Especially in simultaneous speech-to-speech translation \cite{zheng+:2020}, 
there are many efforts have been made
in the simultaneous text-to-text translation stage
to reduce the latency with either fixed \cite{ma+:2019,zheng+opportunistic:2019,zheng+correction:2019} 
or adaptive on-line decoding policy \cite{zheng+:2019,zheng+supervised:2019,zheng+fixed:2019,zheng+:2020}.
But the conventional full-sentence TTS has to wait until the full translation is available, causing the undesirable delay. 
These latencies limit the applicability of neural TTS.

  To reduce these latencies, we propose a neural incremental TTS approach 
  borrowing the recently proposed prefix-to-prefix framework for simultaneous translation \cite{ma+:2019}.
  Our idea is based on two observations: (a)
  in both stages, the dependencies on input are very local (see Fig.~\ref{fig:monoattn}
  for the monotonic attention between text and spectrogram, for example);
  and (b) audio playing is inherently sequential in nature,
  but can be done simultaneously with audio generation, i.e.,
  playing a segment of audio while generating the next.
  In a nutshell,
  we start to generate the spectrogram for the first word after receiving the first two words,
  and this spectrogram is fed into the vocoder right away to generate the waveform for the first word,
  which is also played immediately (see Fig.~\ref{fig:p2p}).
  This results in an $O(1)$ rather than $O(n)$ latency.
Experiments on English and Chinese TTS show that our approach achieves 
similar speech naturalness compared to full sentence methods,
but only with 
a constant (1--2 words) latency.\footnote{
There also exist incremental TTS efforts 
using non-neural techniques \cite{baumann+:2012a,baumann+:2012b,baumann:2014a,pouget+:2015,yanagita+:2018}
which are fundamentally different from our work. See also Sec.~\ref{sec:related}.} 

This paper makes following contributions:
\begin{itemize}
\item From the model point of view, 
with monotonic attention in TTS, we don't need to retrain the model, 
and only need to adapt the inference. This is different from 
all other previous incremental adaptations in simultaneous translation, ASR and TTS \cite{ma+:2019,Novitasari:2019,Yanagita+:2019} 
which rely on new training 
algorithms and/or different training data preprocessing.
\item From a practical point of view, our adaptation reduces the TTS latency from $O(n)$ to $O(1)$, which reduces the TTS response time significantly. 
We also demonstrate that our neural incremental TTS pipeline (including vocoder) can support efficient inference with both CPU and GPU. This is a meaningful step towards the potential use of on-device TTS (as opposed to the prevalent cloud-based TTS).
\end{itemize}

\section{Preliminaries: Neural TTS}
\begin{figure*}[t]
\centering
\includegraphics[width=.99\linewidth]{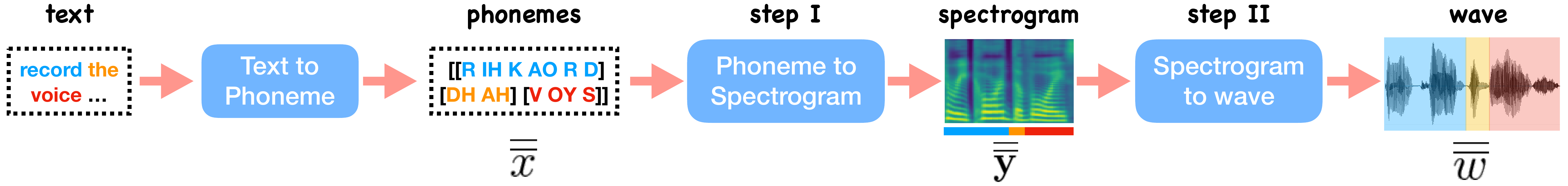}
\caption{Pipeline of conventional full-sentence neural TTS; see also Fig.~\ref{fig:p2p}.}
\label{fig:pipeline}
\end{figure*}

We briefly review the full-sentence 
neural TTS pipeline 
to set up the notations. 
As shown in Fig.~\ref{fig:pipeline}, 
the neural-based text-to-speech synthesis system generally has two main steps:
(1) the text-to-spectrogram step 
which converts a sequence of 
textual features (e.g.~characters, phonemes, words) into another 
sequence of spectrograms (e.g.~mel-spectrogram or linear-spectrogram);
and (2)~the spectrogram-to-wave step, which 
takes the predicted spectrograms 
and generates the audio wave by a  vocoder.

\begin{table}[h]
  \centering
  \resizebox{.48\textwidth}{!}{
\begin{tabular}{ |@{ }c@{ }|c|c|c| } 
 \hline
  & data type & \multicolumn{2}{c|}{examples} \\ 
 \hline
\multirow{3}{*}{\rotatebox{90}{\small subword}} &\multirow{2}{*}{scalar} & $x$ & a phoneme or a char. \\
 &        & $w$ & a sample in wave\\
\cline{2-4}
 & vector & \vecy & a spectrogram frame \\ 
\hline
\hline
\multirow{3}{*}{\rotatebox{90}{\small word}} & \multirow{2}{*}{seq.~of scalars} &  \xbar & phonemes in a word \\
               &      & \wbar &  waveform of a word\\
\cline{2-4}
 & seq.~of vectors & \vecybar & spectrogram of a word \\
\hline
\hline
\multirow{3}{*}{\rotatebox{90}{\small sentence}} & \multirow{2}{*}{seq.~of seq.~of scalars} & \xbarbar
& phonemes in a sentence \\
            &         & \wbarbar  & waveform of a sentence\\
\cline{2-4}
 & seq.~of seq.~of vectors & \raisebox{-.05cm}{$\dbar{\vecy}$} & spectrogram of a sent.\\
\hline
\end{tabular}
}
\caption{Summary of notations. We distinguish vectors (over frequencies) and sequences (over time).}
\label{tab:notations}
\end{table}

\subsection{Step I: Text-to-Spectrogram}

Neural-based text-to-spectrogram systems 
employ the seq-to-seq framework
to encode the source text sequence (characters or phonemes; the latter can be obtained from a prediction model or some heuristic rules;
see details in Sec.~\ref{sec:details})
and decode the spectrogram sequentially 
\cite{wang+:2017,shen+:2018,ping+:2017,li+:2019}.

Regardless the actual design of seq-to-seq framework, 
with the granularity defined on words,
the encoder always takes as input a word sequence 
$\dbar{x}=[\overline{x_1},\overline{x_2}, ... , \overline{x_m}]$
,
where any word $\overline{x_t} = [x_{t,1},x_{t,2},...]$
could be a sequence of phonemes or characters,
and produces another sequence of hidden states 
$
\dbar{\vech}=\dbar{\bm{f}}(\dbar{x}) = \left[\overline{\vech_1},\overline{\vech_2}, ... , \overline{\vech_m}\right] 
$
to represent the textual features (see Tab.~\ref{tab:notations} for notations).

On the other side, the decoder produces the spectrogram $\overline{\vecy_t}$
for the $t^{th}$ word given
the entire sequence of hidden states and the previously generated spectrogram,
denoted by $\dbar{\vecy_{<t}}=[\overline{\vecy_1},...,\overline{\vecy_{t-1}}]$,
where $\overline{\vecy_t} = [\vecy_{t,1},\vecy_{t,2},...]$ is a sequence of spectrogram frames with $\vecy_{t,i} \in \mathbb{R}^{d_y}$
being the $i^{th}$ frame (a vector) of the $t^{th}$ word,
and $d_y$ is the number of bands in the frequency domain (80 in our experiments).
Formally, on a word level, we define the inference process as follows:
\begin{equation}
\overline{\vecy_t} = \overline{\bm{\phi}}(\dbar{x},\dbar{\vecy_{<t}}),
\label{eq:mels2s_word}
\end{equation}
and for each frame within one word, we have
\begin{equation}
\vecy_{t,i} = {\bm{\phi}}(\dbar{x},\dbar{\vecy_{<t}}\circ\dbar{\vecy_{t,<i}}),
\label{eq:mels2s_frame}
\end{equation}
where $\dbar{\vecy_{t,<i}}=[[\vecy_{t,1},...,\vecy_{t,i-1}]]$,
and $\circ$ represents concatenation between two sequences.

\subsection{Step II: Spectrogram-to-Wave}
\label{sec:vocoder}
Given a sequence of acoustic features $\dbar{\vecy}$, the vocoder generates waveform  
$\dbar{w}=[\overline{w_1},\overline{w_2}, ... , \overline{w_m}]$,
where $\overline{w_t} = [w_{t,1},w_{t,2},...]$
is the waveform of the $t^{th}$ word,
given
the linear- or mel-spectrograms. 
The vocoder model  can be either 
autoregressive~\cite{oord+:2016} or 
non-autoregressive ~\cite{oord+:2018, ping+:18, prenger+:19, kim+:2019, yamamoto+:19}.

For the sake of both computation efficiency, and sound quality,
we choose a non-autoregressive model as our vocoder,
which can be defined as follows: without losing generality:
\begin{equation}
\dbar{w} = \dbar{\psi}(\dbar{\vecy},\dbar{z})
\label{eq:vocoder}
\end{equation}
where the vocoder function $\dbar{\psi}$ takes  the
spectrogram $\dbar{\vecy}$ and
a random signal $\dbar{z}$ as input to
generate the wave signal $\dbar{w}$.
Here $\dbar{z}$ is drawn from 
a simple tractable distribution, such as a zero mean spherical Gaussian distribution $\mathcal{N}(0, I)$.
The length of each $\overline{z_t}$ is determined by the length of $\overline{\vecy_t}$,
and we have $|\overline{z_t}| = \gamma \cdot |\overline{\vecy_t}|$.
Based on different STFT procedure, $\gamma$ can be 256 or 300.  
More specifically, the wave generation of the $t^{th}$ word 
can be defined as follows 
\begin{equation}
\overline{w_t} = \overline{\psi}(\dbar{\vecy},\dbar{z}, t)
\label{eq:vocoder_t}
\end{equation}


\section{Incremental TTS}

\if
In above conventional TTS, Text-to-Spectrogram needs to observe full source 
sentence as input to start generating spectrogram,
and the vocoder only enables to generate the wave until all the spectrogram 
representations are fully observed. 
In this section,
we propose a neural-based incremental TTS system to start generating wave
before the full source sentence is observed. 
\fi

Both steps in the above full-sentence TTS pipeline 
require fully observed source text or spectrograms as input.
Here we first propose a general framework to do inference at both
steps with partial source information,
then we present one simple specific example in this framework.

\subsection{Prefix-to-Prefix Framework}
\citet{ma+:2019} propose a prefix-to-prefix framework for simultaneous machine translation. Given a monotonic non-decreasing function $g(t)$, the model would predict each target word $b_t$ based on current available source prefix $\overline{a_{\le g(t)}}$ and the predicted target words $\overline{b_{<t}}$:
\[
p(\overline{b} \mid \overline{a}) = \textstyle\prod_{t=1}^{|\overline{b}|} p(b_t \mid \overline{a_{\leq{g(t)}}},\, \overline{b_{<t}}) 
\]
As a simple example in this framework, they present a wait-$k$ policy, which first wait $k$ source words, and then alternates emitting one target word and receiving one source word. With this policy, the output is always $k$ words behind the input. This policy can be defined with the following function 
\begin{equation}
g_{\text{wait-}k}(t) = \min \{k+t-1, |\overline{a}| \}.
\label{eq:gt-waitk}
\end{equation}

\subsection{Prefix-to-Prefix for TTS}

As shown in Fig.~\ref{fig:monoattn}, there is no long distance reordering between 
input and output sides in the task of Text-to-Spectrogram, 
and the alignment from output side to the input side is monotonic.
One way to utilize this monotonicity is to generate each audio piece for each word independently,
and after generating audios for all words, we can concatenate those audios together. 
However, this naive 
approach mostly produces robotic speech with unnatural prosody.  
In order to generate speech with better prosody,
we need to consider some contextual information 
when generating audio for each word.
This is also necessary to connect audio pieces smoothly. 

To solve the above issue, we propose a  
prefix-to-prefix framework for TTS, which is inspired by the above-mentioned prefix-to-prefix
framework 
for simultaneous translation.  
Within this new framework, our per-word spectrogram $\overline{\vecy_t}$ and wave-form $\overline{w_t}$ 
are both generated incrementally as follows:
\begin{equation}
  \begin{aligned}
    \overline{\vecy_t} &= \overline{\bm{\phi}}(\dbar{x_{\leq g(t)}},\dbar{\vecy_{<t}}),\\    
    \overline{w_t} &= \overline{\psi}(\dbar{\vecy_{\leq h(t)}},\dbar{z_{\leq h(t)}}, t)   
  \end{aligned}
  \label{eq:p2p}
\end{equation}
where $g(t)$ and $h(t)$ are monotonic functions that define
 the number of words being conditioned on 
 when generating results for the $t^{th}$ word.

\subsection{Lookahead-$k$ Policy}
As a simple example in the prefix-to-prefix framework, 
we define two lookahead polices for the two steps (spectrogram and wave)
with $h(\cdot)$ and $g(\cdot)$ functions, resp.
These are similar to the monotonic function in wait-$k$ policy \cite{ma+:2019}
in Eq.~\ref{eq:gt-waitk} (except that lookahead-$k$ is wait-$(k\!+\!1)$):
\begin{equation}
\begin{matrix}
g_{\text{lookahead-}k_1}(t) = \min\{k_1 + t , |\dbar{x}|\} \\
h_{\text{lookahead-}k_2}(t) = \min\{k_2 + t , |\dbar{\vecy}|\} 
\end{matrix}
\label{eq:waitk}
\end{equation}

Intuitively, the function $g_{\text{lookahead-}k_1}(\cdot)$ implies that the spectrogram generation of the $t^{th}$ word
is conditioned on $(t+k_1)$ words, with the last $k_1$ being the lookahead. 
Similarly, the function $h_{\text{lookahead-}k_2}(\cdot)$ implies that the wave generation of the $t^{th}$ word
is conditioned on $(t+k_2)$ words' spectrograms. 
Combining these together, we can obtain a lookahead-$k$ policy for the whole TTS system, where $k=k_1+k_2$.
An example of lookahead-1 policy is provided in Fig.~\ref{fig:p2p}, where we take $k_1=1$ for the spectrogram generation and $k_2=0$ for the wave generation. 


\section{Implementation Details}
\label{sec:details}
In this section, we provide some implementation details for the two steps (spectrogram and wave).
We assume the given text input is normalized, and we use an existing grapheme-to-phoneme tool\footnote{We use g2pE~\cite{g2pE2019} (\url{https://github.com/Kyubyong/g2p}) for English and pypinyin (\url{https://github.com/mozillazg/python-pinyin}) for Chinese. } to generate phonemes for the given text. For some languages like Chinese, we need to use an existing tool\footnote{We use jieba (\url{https://github.com/fxsjy/jieba}) to do text segmentation for Chinese.} to do text segmentation before generating phonemes.

In the following, we assume the pre-trained models 
for both steps are given, and we only perform inference-time adaptations.
For the first step, we use the Tacotron 2 model~\cite{shen+:2018}, which takes generated phonemes as input,
and for the second step we use the Parallel WaveGAN vocoder~\cite{yamamoto+:19}.


\subsection{Incremental Generation of Spectrogram}

Different from full sentence scenario,
where we feed the entire source text to the 
encoder,
we gradually provide source text input to the model word by word 
when more input words are available.
By our prefix-to-prefix framework, we will predict mel spectrogram for 
the $t^{th}$ word, when there are $g(t)$ words available. 
Thus, the decoder predicts the $i^{th}$ spectrogram frame of the
$t^{th}$ word with only partial source information as follows:
\begin{equation}
\vecy_{t,i} = \bm{\phi}(\dbar{x_{\leq g(t)}},\dbar{\vecy_{< t}}\circ \dbar{\vecy_{t,<i}})
\label{eq:decoder}
\end{equation}
where $\dbar{\vecy_{t,<i}}=[[\vecy_{t,1},...,\vecy_{t,i-1}]]$ represents
the first $i-1$ spectrogram frames in the $t^{th}$ word.

In order to obtain the corresponding relationship between
the predicted spectrogram and the currently available source text,
we rely on the attention alignment applied in our decoder, which 
is usually monotonic.  
To the $i^{th}$ spectrogram frame of the $t^{th}$ word, 
we can define the attention function $\bm{\sigma}$ in our decoder as follows
\begin{equation}
{\vecc}_{t,i} = \bm{\sigma}(\dbar{x_{\leq t+1}},\dbar{\vecy_{< t}}\circ \dbar{\vecy_{t,<i}})
\label{eq:attention}
\end{equation}
The output $\vecc_{t,i}$
represents the alignment distribution over the input text 
for the $i^{th}$ predicted spectrogram frame.  
And we choose the input element with the highest probability 
as the corresponding input element for this
predicted spectrogram, that is, $\argmax \vecc_{t,i}$.
When we have $\argmax \vecc_{t,i} > \sum_{\tau=1}^{t}|\overline{x_{\tau}}|$, 
it implies that the $i^{th}$ spectrogram frame corresponds to the $(t+1)^{th}$ word,
and all the spectrogram frames for the $t^{th}$ word are predicted.


When the encoder observes the entire source sentence,
a special symbol \eos was feed into the encoder, 
and the decoder continue to generate spectrogram word by word.
The decoding process ends when the binary ``stop'' predictor of the model  
predicts the probability larger than $0.5$.

\subsection{Generation of Waveform}\label{sec:gen_wav}

After we obtain the predicted spectrograms for a new word, we feed them into 
our vocoder to generate waveform.  
Since we use a non-autoregressive vocoder, 
we can generate each audio piece for those given spectrograms 
in the same way as full sentence generation. Thus, 
we do not need to make modification on the vocoder model implementation. 
Then the straightforward way to generate each audio piece is to apply
Eq.~\ref{eq:vocoder_t} at each step $t$ conditioned on the spectrograms
of each word $\overline{\vecy_t}$.  
However, when we concatenate the audio pieces generated in this way, 
we observe some noise at the connecting part of two audio pieces.

To avoid such noise, we 
sample a long enough random vector as the input vector $\dbar{z}$
and fix it when generating audio pieces.
Further, we append additional $\delta$ number of spectrogram frames to each side of the current spectrograms $\overline{\vecy_t}$ if possible.
That is, at most $\delta$ number of last frames in $\overline{\vecy_{t-1}}$ are added in front of $\overline{\vecy_t}$, and at most $\delta$ number of first frames in $\overline{\vecy_{t+1}}$ are added at the end of $\overline{\vecy_t}$.
This may give a longer audio piece than we need, so we can remove the extra parts from that.
Formally, the generation procedure of wave for each word can be defined as follows
\begin{equation}
\overline{w_{t}} = \overline{\psi}(\dbar{\vecy_{[t-1:h(t)]}},\dbar{z_{[t-1:h(t)]}}, t)
\label{eq:overlap}
\end{equation}
where $\dbar{\vecy_{[t-1:h(t)]}} = [\overline{\vecy_{t-1}}, \dots, \overline{\vecy_{h(t)}}]$ and 
$\dbar{z_{[t-1:h(t)]}} = [\overline{z_{t-1}}, \dots, \overline{z_{h(t)}}]$. 


\if
That is, when generating the audio piece $\overline{w_t}$ 
ranged from $a$ to $b$ in the full sentence audio $\vecw$,  the chosen input vector $z_t$ will also correspond to the same range in vector $\vecz$.  
Further, for a given mel segment $y_t$, we append at most $\delta$ number of mel frames to its each side if possible, where $\delta$ is a hyper-parameter. That is, at step $t$, we are given a sequence of mel segments $\vecy_{\le h(t)}$, and we append the last $\alpha = \min\{ \delta, |y_{t-1}| \}$ number of mel frames from $y_{t-1}$ at the beginning of $y_t$ if $t>1$, and append the first $\beta = \min \{\delta, |y_{t+1}| \}$ number of mel frames of $y_{t+1}$ at the end of $y_t$ if this is not the last step. 
The generated audio piece $w'_t$ from this extended mel segment will overlap with the previous audio piece $w_{t-1}$ and the next audio piece $w_{t+1}$.
So we need to remove the extra parts from $w'_t$, which correspond to those appended mel frames according to the used sample rate. 
In this way, we can almost avoid the connection noise.
\fi

\section{Related Work}

\label{sec:related}

\if
Incremental TTS is previously studied in the statistical parametric speech synthesis framework based on Hidden Markov Model (HMM).
Such kind of framework usually consists of several steps: extracting linguistic features, establishing an HMM sequence to estimate acoustic features, constructing speech waveform from those acoustic features.  
Based on this framework,~\citet{baumann+:2012a} propose an incremental spoken dialogue system architecture and toolkit called {\em INPROTK}, including recognition, dialogue management and TTS modules. 
With this toolkit,~\citet{baumann+:2012b} present a component for incremental speech synthesis, which is not fully incremental on the HMM level.
\citet{pouget+:2015} propose a training strategy based on HMM with unknown linguistic features for incremental TTS.
\citet{baumann:2014b, baumann:2014a} proposes to flexibly use linguistic features and choose default values when they are not available.
The above works all focus on stress-timed languages, such as English and German, while~\citet{yanagita+:2018} propose a system for Japanese, a mora-timed language.
Although these works show speech quality can be improved for incremental TTS, these systems require full context labels of linguistic features, making it difficult to improve the audio quality when input text is revealed incrementally.
Further, each component in their systems is trained and tuned separately, resulting in error propagation.

Neural speech synthesis systems provide a solution for this problem.
Such systems do not need the full context labels of linguistic features any more, 
and the quality of synthesized speech with those systems have obtained the start-of-the-art results. 
Several different systems are proposed, including Deep Voice~\cite{arik+:2017}, Deep Voice 2~\cite{gibiansky+:2017}, Deep Voice 3~\cite{ping+:2017}, Tacotron~\cite{wang+:2017}, Tacotron 2~\cite{shen+:2018}, ClariNet~\cite{ping+:18}.
However, these systems all need the entire sentence as input to generate the speech, resulting in large latency for some applications such as spoken dialogue system and speech simultaneous translation system.
More recently, some parallel systems are proposed for TTS~\cite{ren+:2019, peng+:2019}, which avoid the autoregressive steps and provide faster audio generation. 
But these systems still suffer from large input latency compared with real incremental TTS system, since they generate waveform on a sentence-level instead of word-level, implying that they will need to wait for long enough input to start speech generation.
Our proposal is a general adaptation that also can be easily applied to the above parallel systems to further improve the generation efficiency 
in the full-sentence generation scenario.
\fi

There are some existing work about incremental TTS based Hidden Markov Model (HMM).
\citet{baumann+:2012a} propose an incremental spoken dialogue system architecture 
and toolkit called {\em INPROTK}, including recognition, dialogue management and TTS modules. 
With this toolkit, ~\citet{baumann+:2012b} present a component for incremental speech synthesis, 
which is not fully incremental on the HMM level.
\citet{pouget+:2015} propose a training strategy based on HMM with unknown 
linguistic features for incremental TTS.
\citet{baumann:2014b, baumann:2014a} proposes use linguistic features 
and choose default values when they are not available.
The above works all focus on stress-timed languages, such as English and German, 
while~\citet{yanagita+:2018} propose a system for Japanese, a mora-timed language.
These systems require full context labels of linguistic features, 
making it difficult to improve the audio quality when input text is revealed incrementally.
Further, each component in their systems is trained and tuned separately,
resulting in error propagation.

There is parallel work from \namecite{Yanagita+:2019}, which
introduced a different neural approach for segment-based incremental TTS. 
Their proposed solution synthesizes each segment (could be as long as half sentence) at a time, 
thus not strictly incremental on the word level. 
When they perform word-level synthesis, as it is shown in their paper, 
there is a huge performance drop from 3.01 (full-sentence) to 2.08.
Their proposed approach has to retain the basic full-sentence model with segmented texts 
and audios which were obtained from forced alignment (different models for different latencies),
while we only make adaptations to the decoder at inference time with an existing 
well-trained full-sentence model.
Our model not only uses previous context, but also use limited, 
a few lookahead words for better prosody
and pronunciation. 
The above advantages of our model guarantee that our model achieves 
similar performance with full-sentence model 
with much lower latency on word-level inference.
On the contrary, the model from \namecite{Yanagita+:2019}
did not use lookahead information at all, which can be problematic in the cases
when word has multiple pronunciation that depends on following word.
For example, there are two pronunciations for the word ``the'' which are 
``DH IY'' and ``DH AH''.
When the word after ``the'' starts with vowel sound, ``DH IY'' is the correct option
while ``DH AH'' is used only when the following word begins with consonant sound. 
Lookabead information is more important in liaison, 
where the final consonant of one word links with the first vowel of the next word, e.g.,
``an apple'', ``think about it'', and ``there is a''.
This problem is even more severe in other languages like French. More generally, co-articulation is common in most languages, where lookahead is needed.

\section{Experiments}

\subsection{Experimental Setup}

\begin{table*}[h]
\centering
\resizebox{2\columnwidth}{!}{%
\begin{tabular}{|c|c|c|c|c|c|c|}
\hline
 & \multicolumn{3}{c|}{English} & \multicolumn{3}{c|}{Chinese} \\ \hline
 Methods&
  MOS $\uparrow$&
  \begin{tabular}[c]{@{}c@{}}duration \\ deviation  \end{tabular}$\downarrow$ &
  \begin{tabular}[c]{@{}c@{}}pitch \\ deviation \end{tabular} $\downarrow$&
  MOS $\uparrow$&
  \begin{tabular}[c]{@{}c@{}}duration \\ deviation \end{tabular} $\downarrow$ &
  \begin{tabular}[c]{@{}c@{}}pitch \\ deviation \end{tabular} $\downarrow$ \\ \hline
Ground Truth Audio & $ 4.40 \pm 0.04$     & -        & -       & $ 4.37 \pm 0.04$    & -        & -       \\ \hline
Ground Truth Mel   & $4.25 \pm 0.04$    & -        &         & $4.35 \pm 0.04$    & -        & -       \\ \hline
Full-sentence      & $4.20 \pm 0.05$    & -        & -       & $4.28 \pm 0.04$    & -        & -       \\ \hline
\hline
Lookahead-2 ($k_1\!=\!1, k_2\!=\!1$) $\dagger$ & $4.19 \pm 0.05$    & 14.05    & 18.69   & $4.22 \pm 0.04$    & 23.97    & 21.42   \\ \hline
Lookahead-1 ($k_1\!=\!1, k_2\!=\!0$) $\dagger$ & $4.18 \pm 0.05$    & 14.79    & 19.55   & $4.18 \pm 0.04$   & 24.11    & 21.15   \\ \hline
Lookahead-0 ($k_1\!=\!0, k_2\!=\!0$) $\dagger$ & $3.74 \pm 0.06$   & 35.93    & 33.51   & $4.09 \pm 0.04$    & 27.09    & 28.06   \\ \hline
\hline
\citet{Yanagita+:2019} (2 word)     & $3.99 \pm 0.06$    & 29.09    & 35.63   & -       & -        & -       \\ \hline
\citet{Yanagita+:2019} (1 word)     & $3.76 \pm 0.07$    & 36.13    & 40.26   & -       & -        & -       \\ \hline
\citet{Yanagita+:2019} (lookahead-0)      & $3.89 \pm 0.06$    & 29.08    & 37.12   & -       & -        & -       \\ \hline
Lookahead-0-indep     & $2.94 \pm 0.09$   & 101.01   & 48.51   & $2.50 \pm 0.05$   & 64.52    & 50.28   \\ \hline
\end{tabular}
}
\caption{
    \textbf{MOS ratings:} with 95\% confidence intervals for comparing the audio qualities of different methods on English and Chinese. 
    We can incrementally synthesize high quality audios with our lookahead-1 and lookahead-2 policies. 
    The method of~\citet{Yanagita+:2019} uses augmented data to train the model and needs more steps to converge, but its audio quality is worse than that of lookahead-1 policy. \textbf{Prosody analysis:} phoneme level duration (in ms) and pitch deviation (in Hz) RMSE of different methods compare against to full-sentence (smaller RMSE is better) in English and Chinese. In full-sentence generation of English, the mean phoneme duration and pitch are 97.41 ms and 237.23 Hz respectively. In full-sentence generation of Chinese, the mean phoneme duration and pitch are 89.93 ms and 252.73 Hz respectively. $\dagger$ represents the performance of our proposed methods.
    }

\label{tab:bigmos}
\end{table*}

\paragraph{Datasets}
We evaluate our methods on English and Chinese. 
For English, we use a proprietary speech dataset containing 13,708 audio clips (i.e., sentences) from a female speaker and the corresponding transcripts.
For Chinese, we use a public speech dataset\footnote{\url{https://www.data-baker.com/open_source.html}} containing 10,000 audio clips from a female speaker and the transcripts.
We downsample the audio data to 24 kHz,
and split the dataset into three sets: the last 100 sentences for testing, the second last 100 for validation and the others for training.
Our mel-spectrogram has 80 bands, and is computed through a short time Fourier transform (STFT) with window size 1200 and hop size 300.

\paragraph{Models}
We take the Tacotron 2 model~\cite{shen+:2018} as our phoneme-to-spectrogram model and train it with additional {\em guided attention loss}~\cite{tachibana+:2018} which speeds up convergence.
Our vocoder is the same as that in the Parallel WaveGAN paper~\cite{yamamoto+:19}, which consists of 30 layers of dilated residual convolution blocks with exponentially increasing three dilation cycles, 64 residual and skip channels and the convolution filter size 3.  


\paragraph{Inference}
In our experiments, we find that synthesis on a word-level severely slows down synthesis,
because many words are synthesized more than once due to overlap
(our method will generate at most $2\delta$ additional spectrogram frames for each given spectrogram sequence, as described in Sec.~\ref{sec:gen_wav}). 
Therefore, below we do inference on a {\em chunk-level},
where each chunk consists of one or more words depending on a hyper-parameter $l$:
a chunk contains the minimum number of words such that the number of phonemes in this chunk is at least $l$ which is 6 for English and 4 for Chinese.

In the following sections, we 
consider three different policies: 
lookahead-2 ($k_1\!=\!1$ in text-to-spectrogram, $k_2\!=\!1$ in spectrogram-to-wave), lookahead-1 policy ($k_1\!=\!1, k_2\!=\!0$) and lookahead-0 policy ($k_1\!=\!k_2\!=\!0$).
For lookahead-2 policy, we set $\delta=15$ on English and $\delta=10$ on Chinese (see Sec.~\ref{sec:gen_wav} for the definition of $\delta$).
All methods are with GeForce TITAN-X GPU.



\subsection{Audio Quality}

In this section, we compare the audio qualities of different methods.
For this purpose, we choose 80 sentences 
 from our test set and generate audio samples for these sentences with different methods, which include (1) {\em Ground Truth Audio}; 
(2) {\em Ground Truth Mel}, where we convert the ground truth mel spectrograms into audio samples using our vocoder; 
(3) {\em Full-sentence}, where we first predict all mel spectrograms given the full sentence text and then convert those to audio samples; 
(4) {\em Lookahead-2}, where we incrementally generate audio samples with lookahead-2 policy; 
(5) {\em Lookahead-1}, where we incrementally generate audio samples with lookahead-1 policy; 
(6) {\em Lookahead-0}, where we incrementally generate audio samples with lookahead-0 policy; 
(7) {\em \citet{Yanagita+:2019} (2 words)}, where we follow the method in~\citet{Yanagita+:2019} and synthesize with incremental unit as two words;
(8) {\em \citet{Yanagita+:2019} (1 word)}, where we follow the method in~\citet{Yanagita+:2019} and synthesize with incremental unit as one word\footnote{We use the ``Independent'' approach from the original paper for connecting the units in the both baselines, as this is shown to have better prosody in the original paper~\cite{Yanagita+:2019}.};
(9) {\em Lookahead-0-indep}, where we generate audio pieces independently for each chunk without surrounding context information.   
These audios are sent to Amazon Mechanical Turk where each sample received 10 human ratings scaled from 1 to 5. 
The MOS (Mean Opinion Score) of this evaluation is provided in Table~\ref{tab:bigmos}.


From Table~\ref{tab:bigmos}, we notice that lookahead-2 policy generates 
comparable audio quality to the full-sentence method. 
Lookahead-0 has poor performance due to lack of following words' information.
But it still outperforms lookahead-0-indep since lookahead-0-indep does not 
use any previous context information.
Note that we use a neural vocoder to synthesize our audios in the two~\citet{Yanagita+:2019} baselines,
and their MOS scores in the above table are much higher than then original paper.

Following the prosody analysis in \cite{Baumann:2012}, 
we perform the similar prosody analysis of the 
difference between various methods in Table ~\ref{tab:bigmos}.
Duration and pitch are two essential components for prosody.
We evaluate how the duration and pitch under different incremental generation 
settings deviate from those in full-sentence with root mean squared error (RMSE).


The RMSE for both 
duration and pitch of lookahead-1 and lookahead-2
are much lower compared with lookahead-0-indep and lookahead-0. 
The RMSE of lookahead-2 is slightly better than lookahead-1 which also agrees the 
results of MOS in Table ~\ref{tab:bigmos}. 
Compared with~\citet{Yanagita+:2019}'s models, 
 lookahead-1 and lookahead-2 achieves much better duration and pitch
RMSE.

In the cases of lookahead-0, our proposed model is slightly worse (0.15 in MOS, about 3.8$\%$) than~\citet{Yanagita+:2019}'s models since we don't retrain the model. But \citet{Yanagita+:2019}'s model needs retraining and special preprocessing of training data. In all other settings, lookahead-1 and lookahead-2, our model gets the best performance.

As discussed in latter part of Section.\ref{sec:related}, some languages seem to require less lookahead; for example, our experiments on Chinese TTS in this paper showed that improvement from lookahead is smaller than English in Table ~\ref{tab:bigmos}. However, this is due to the fact that our Chinese dataset is mostly formal text that does not expose co-articulation, but in informal fast speech, co-articulation between word boundaries is more common (such as third-tone sandhi)
where you need lookahead \cite{chen+yuan:2007,chen+yuan:2014}.

\subsection{Visual Analysis}

\begin{figure}[h]
\centering
\includegraphics[width=\linewidth]{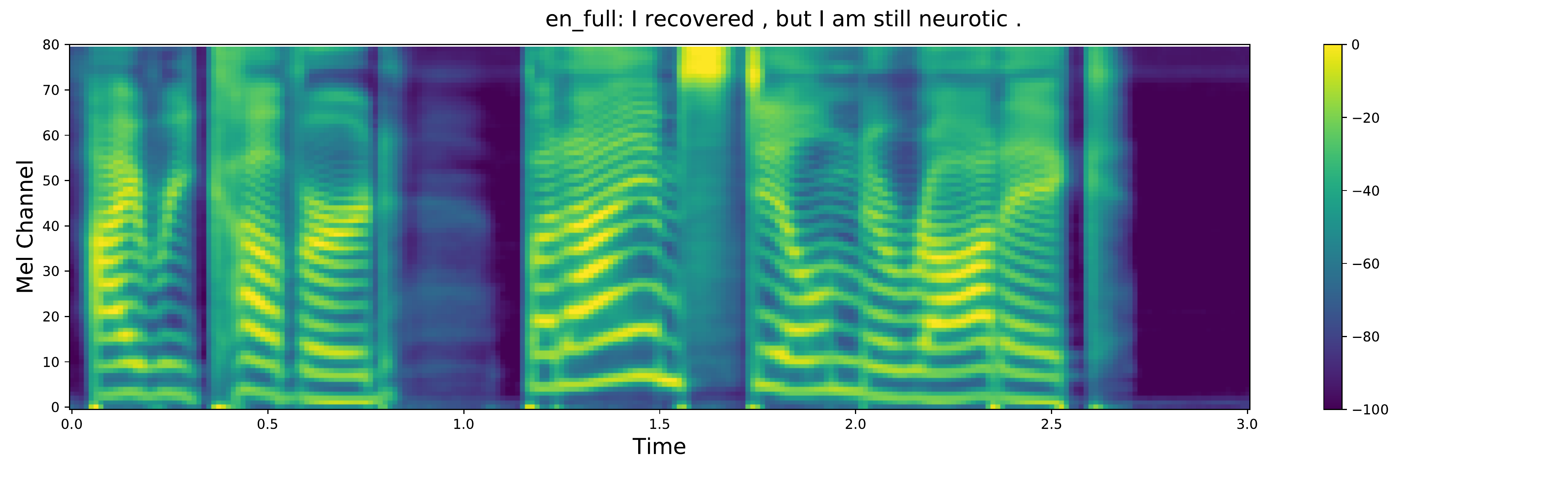}
\includegraphics[width=\linewidth]{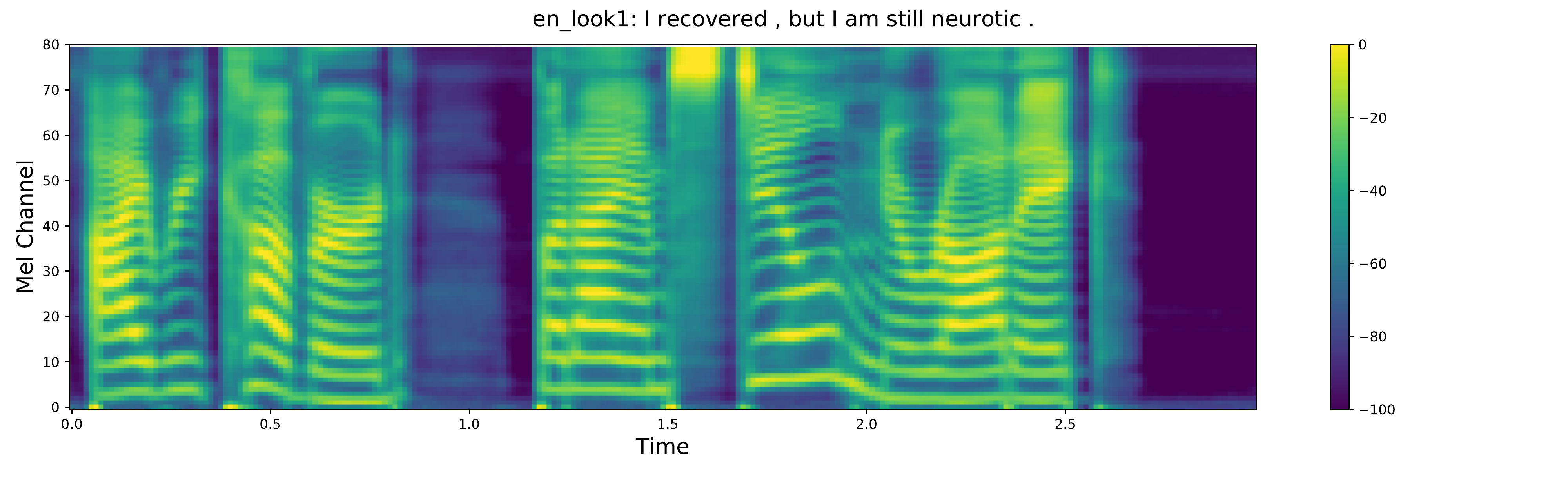}
\caption{Mel-spectrograms comparison between full-sentence TTS (top), and our lookahead-1 policy (bottom). These two mel-spectrograms are very similar.
}
\label{fig:mel}
\end{figure}

To make visual comparison, Fig.~\ref{fig:mel} shows mel-spectrograms obtained from
full-sentence TTS and lookahead-1 policy. 
We can see that the mel-spectrogram from lookahead-1 policy is very similar to that by full-sentence TTS.
This comparison also proves that our incremental TTS
can approximate the quality of full-sentence TTS.

\subsection{Latency}

\begin{figure}[t]
\centering
\includegraphics[width=\linewidth]{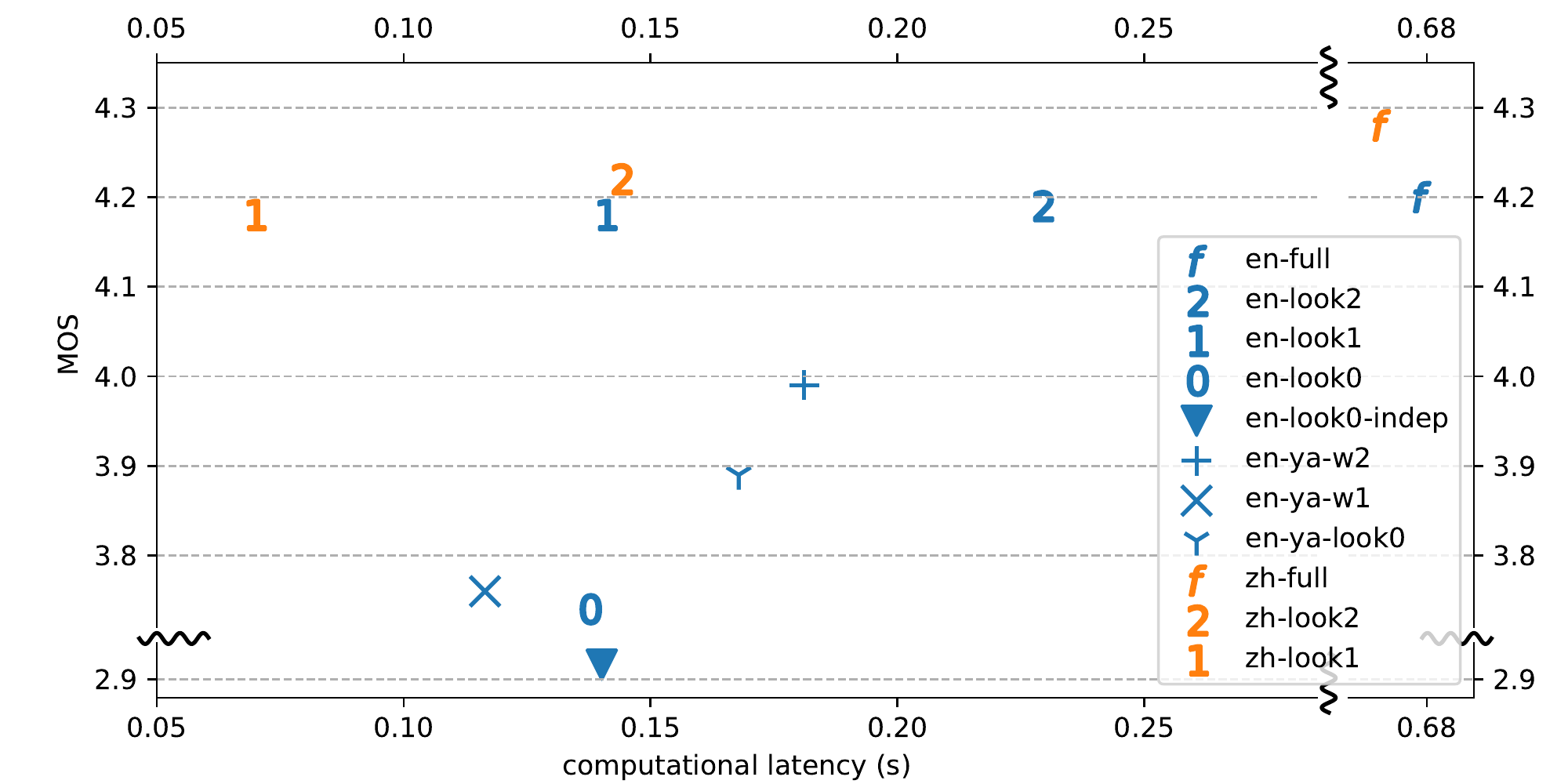}
\caption{
    MOS score against computational latency for English and Chinese. ``look''$*$ denotes lookahead-$*$, and ``ya'' denotes baselines from \citet{Yanagita+:2019}.
}
\label{fig:mos-lat}
\end{figure}

We next compare the latencies of full-sentence TTS and our proposed lookahead-2 and lookahead-1 policies. 
We consider two different scenarios: (1) when all text input is immediately available; and (2) when the text input is revealed incrementally.
The first setting is the same as conventional TTS, while the second is required in
applications like
simultaneous translation, dialog generation, and assistive technologies.


\subsubsection{All Input Available}
\begin{figure}[!t]
\centering
\includegraphics[width=\linewidth]{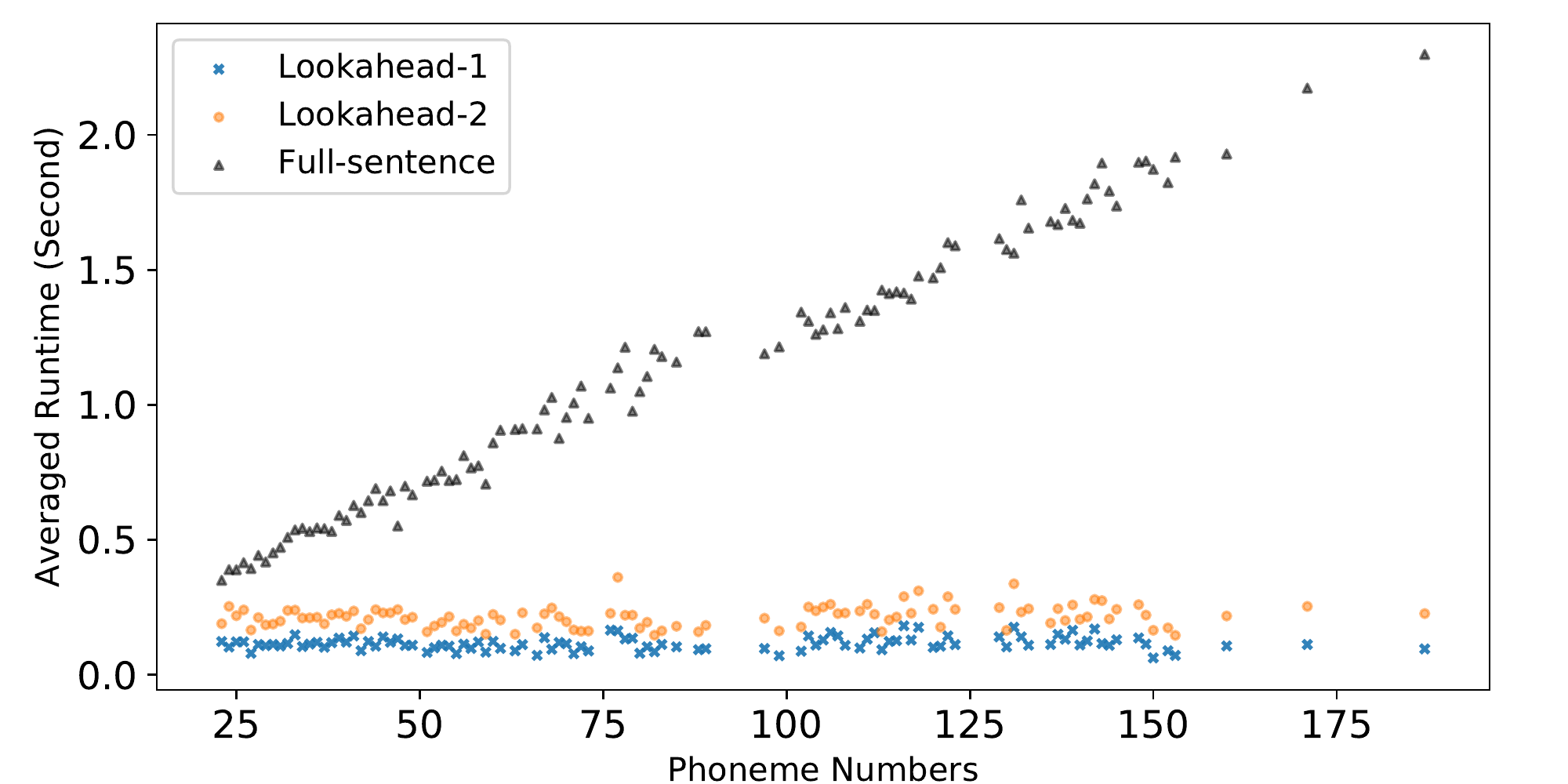}
\includegraphics[width=\linewidth]{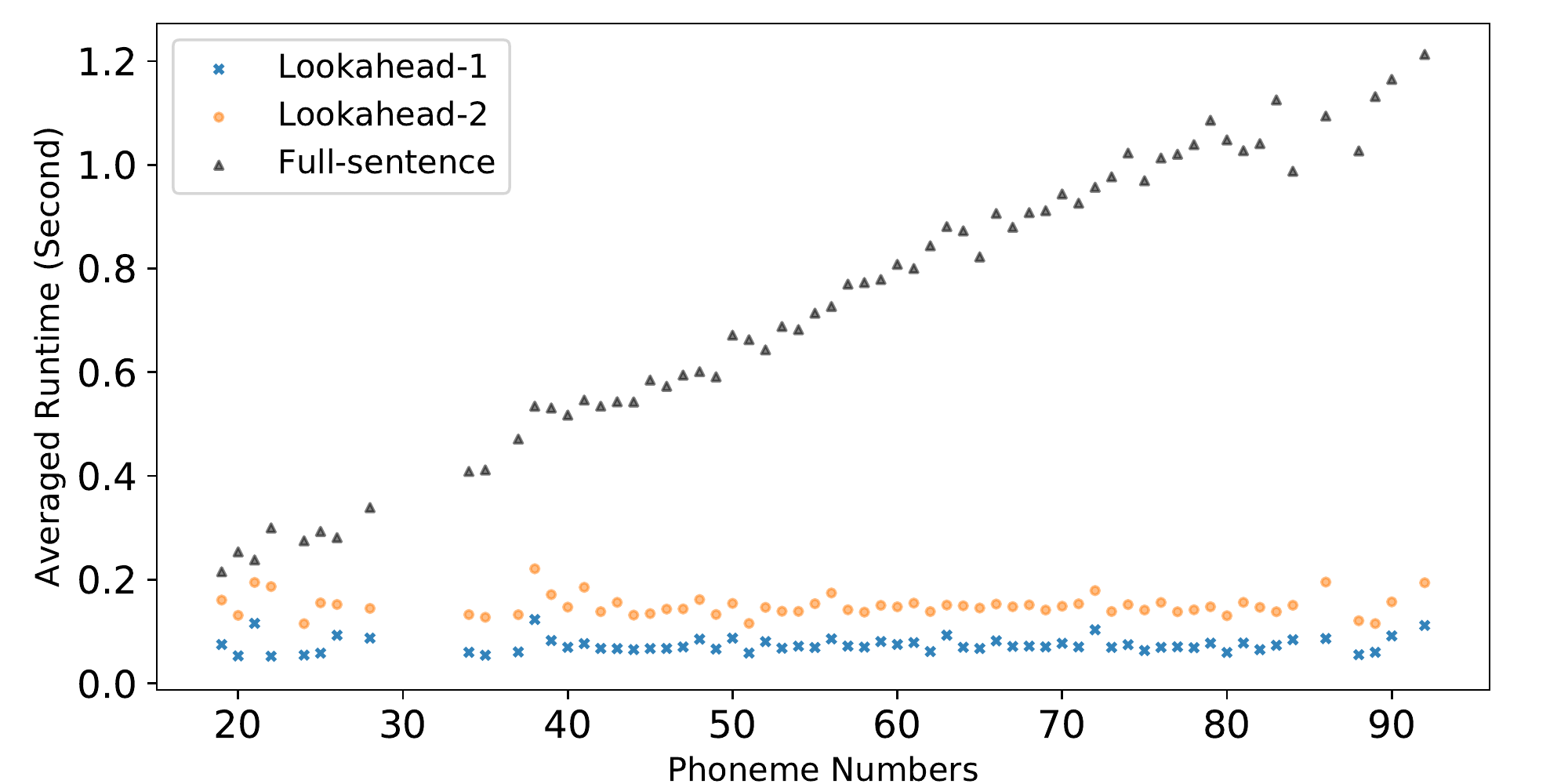}
\caption{Averaged computational latency of different methods for English (upper) and Chinese (lower). 
Full-sentence method has its latency increasing with the sentence length, while our incremental methods have constant latency with different sentence lengths.
}
\label{fig:gens}
\end{figure}



\begin{figure}[t]
\centering
\includegraphics[width=0.49\linewidth]{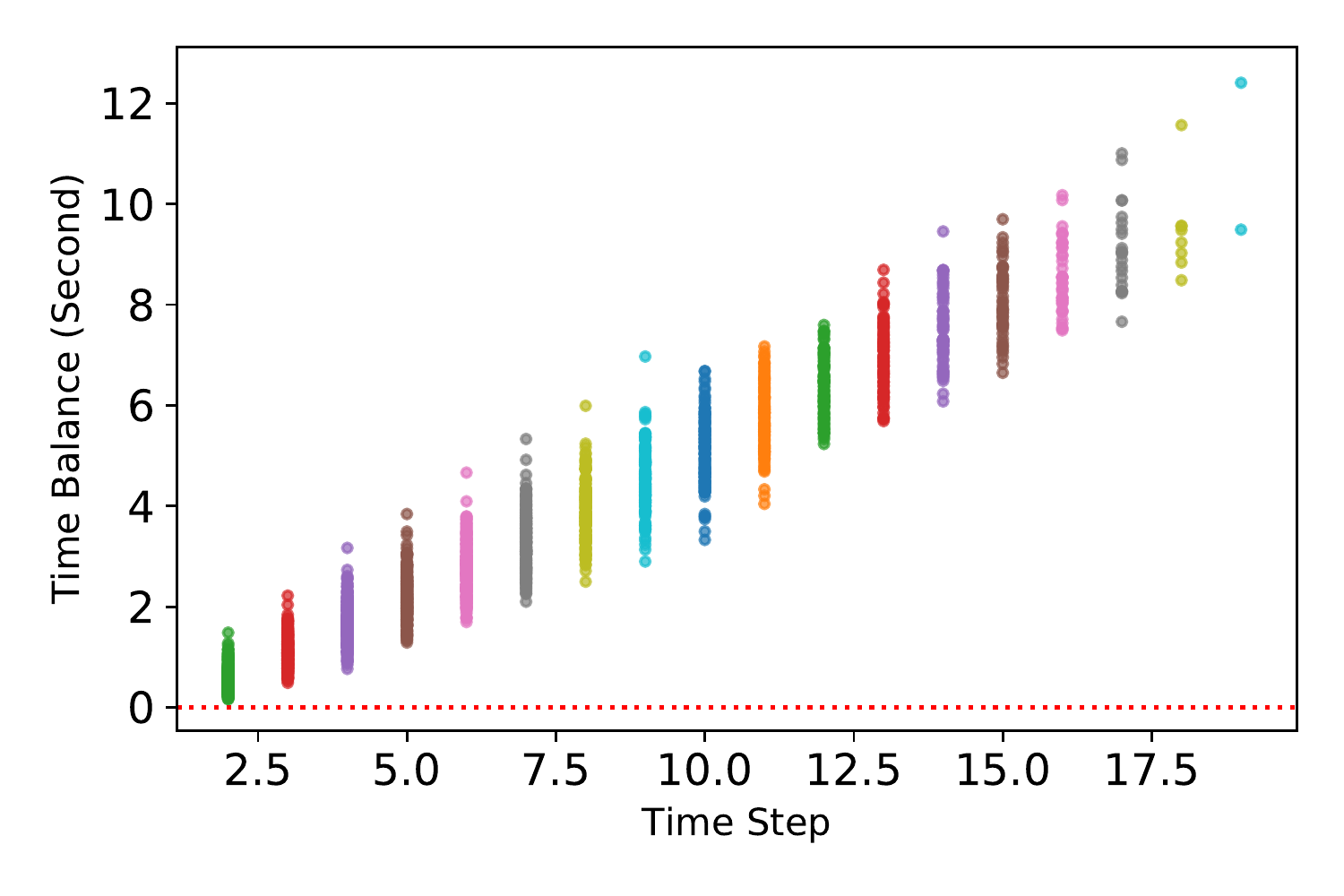}
\includegraphics[width=0.49\linewidth]{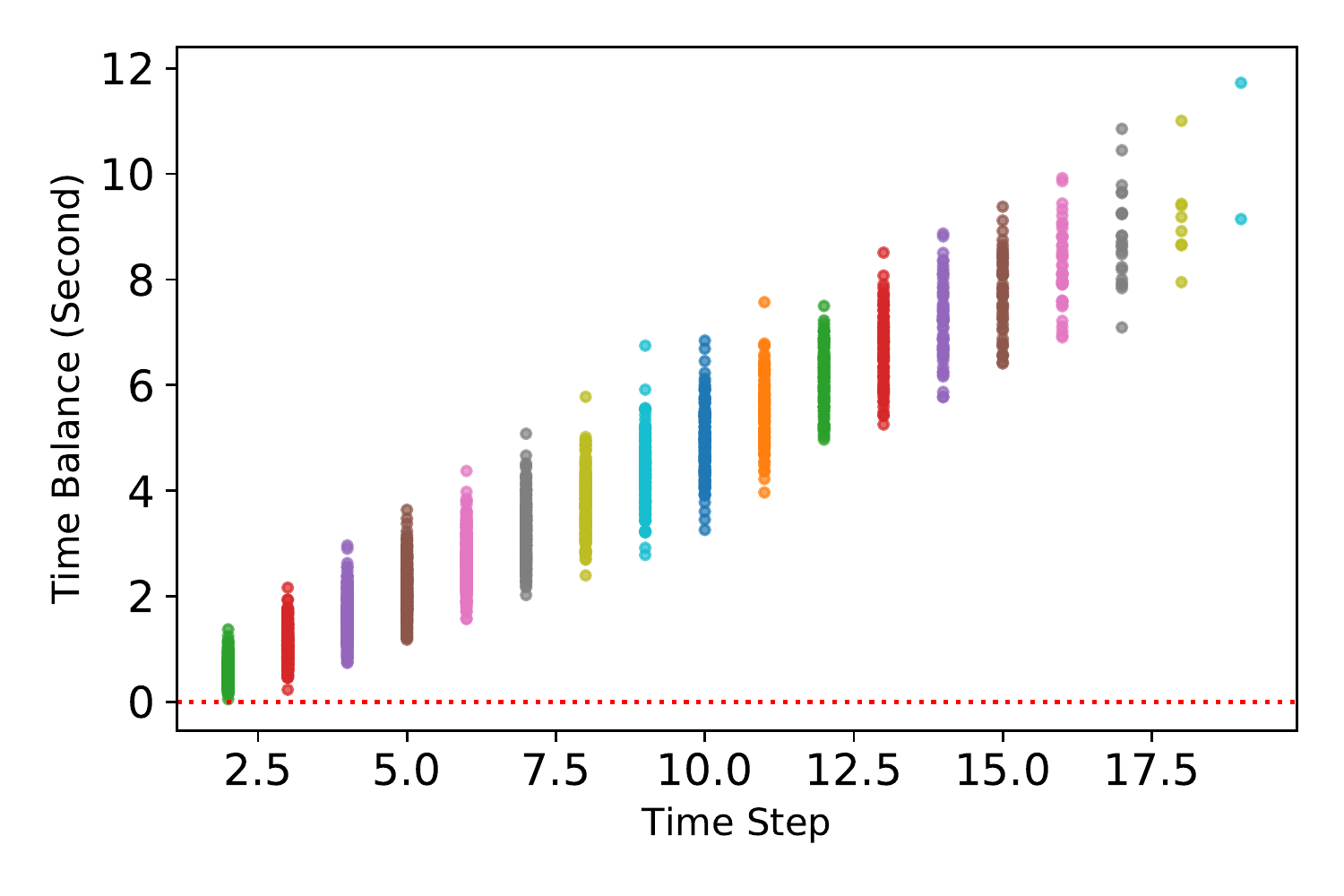}
\includegraphics[width=0.49\linewidth]{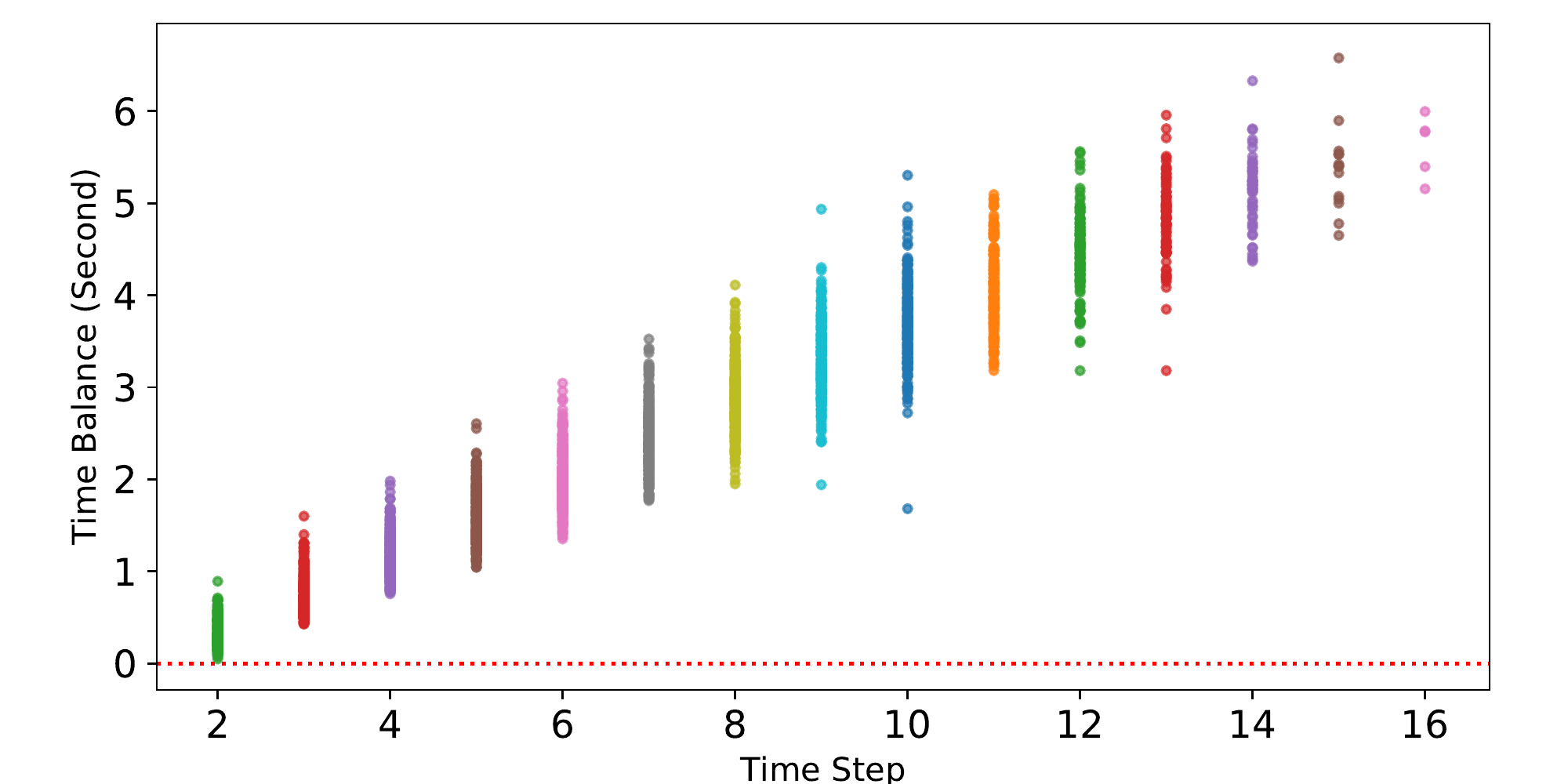}
\includegraphics[width=0.49\linewidth]{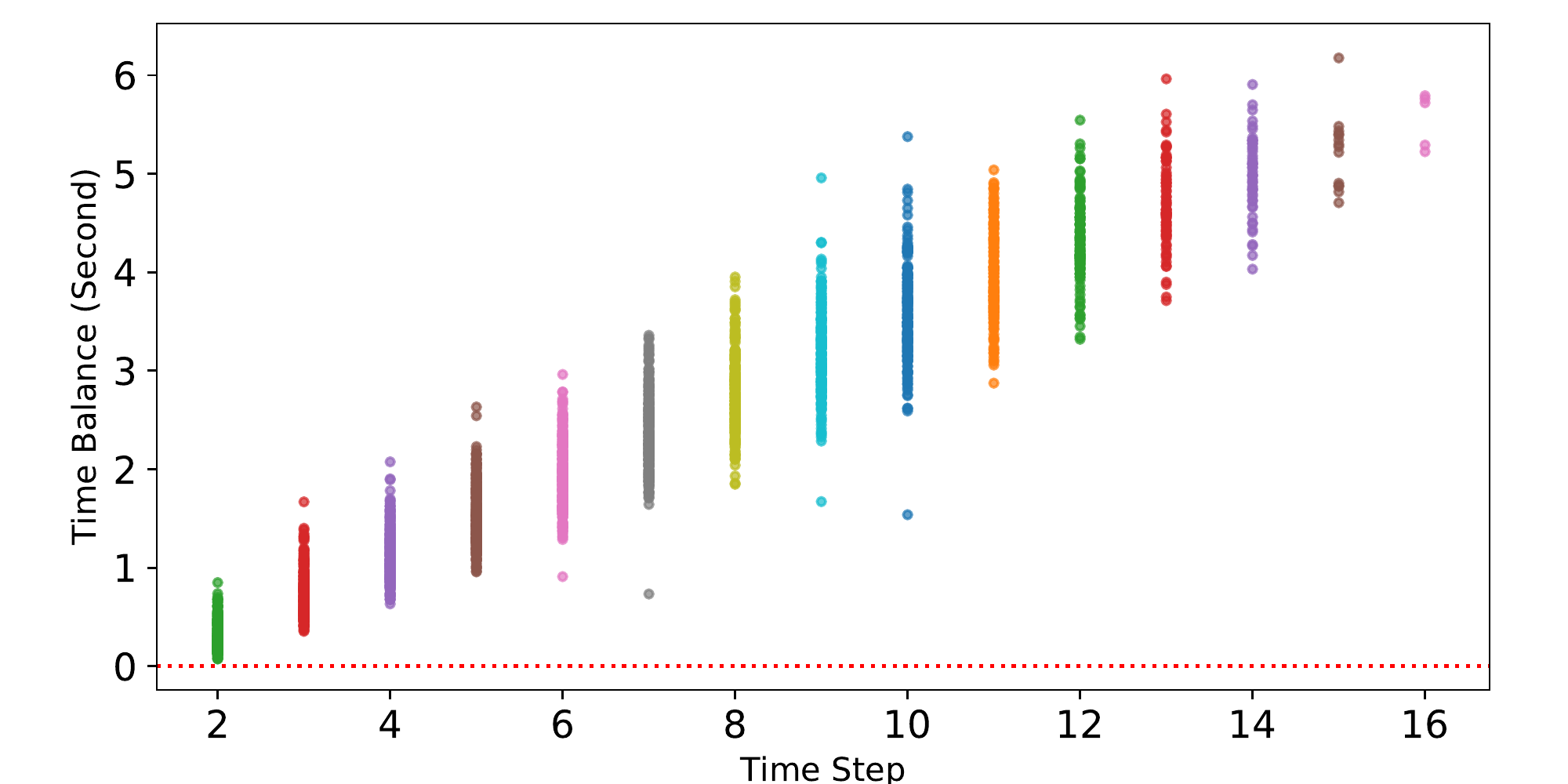}
\caption{
    Time balance $\tb(t)$ (the higher the better) 
  for all sentences in the 300-sentence set for English (upper) and Chinese (lower).
  Audio can play continuously if $\tb(t)\ge 0$ for all $t$, which is true for all plots. 
  Lookahead-1 policy is on the left side and lookahead-2 is on the right side.
    }
\label{fig:diff}
\end{figure}


For this scenario, there is no input latency, and we only need to consider computational latency.
For full-sentence method, this will be the synthesizing time of the whole audio sample; while for our incremental method, this latency will be the synthesizing time of the first chunk if the next audio piece can be generated before the current audio piece playing finishes.  
We first compare this latency, and then show the audio pieces can be played continuously without interruptions.
Specifically, we do inference with different methods on 300 sentences (including 100 sentences from our validation set, test set and training set respectively) and average the results over sentences with the same length.
The results for English and Chinese are provided in Fig.~\ref{fig:gens}.

As shown in Fig.~\ref{fig:gens}, 
we observe that the latency of full-sentence TTS  scales linearly with sentence length,
being 1.5+ seconds for long English sentences (125+ phonemes) and 
1+ seconds for long Chinese sentences (70+ phonemes).
By contrast, our incremental TTS have constant latency that does {\em not} grow with  sentence length,
which is generally under 0.3 seconds for both English and Chinese regardless of different sentence length.


Fig.~\ref{fig:mos-lat} compares the latency and MOS with different policies against to
several baselines from \citet{Yanagita+:2019} on English dataset.
To make a fair comparison with baseline, we use the model from \citet{Yanagita+:2019}
and follow our lookahead-0 policy to generate ``en-ya-look0'' in Fig.~\ref{fig:mos-lat}.
Compared with lookahead-0, ``en-ya-look0'' has higher MOS score since it is retrained with
chunk-based dataset.
However, when a small amount of lookahead is allowed, our lookahead 1 and 2 outperform
``en-ya-w1'' and ``en-ya-w2'' easily.
This also demonstrate the importance of lookahead information.

\begin{figure}[t]
\centering
\includegraphics[width=\linewidth]{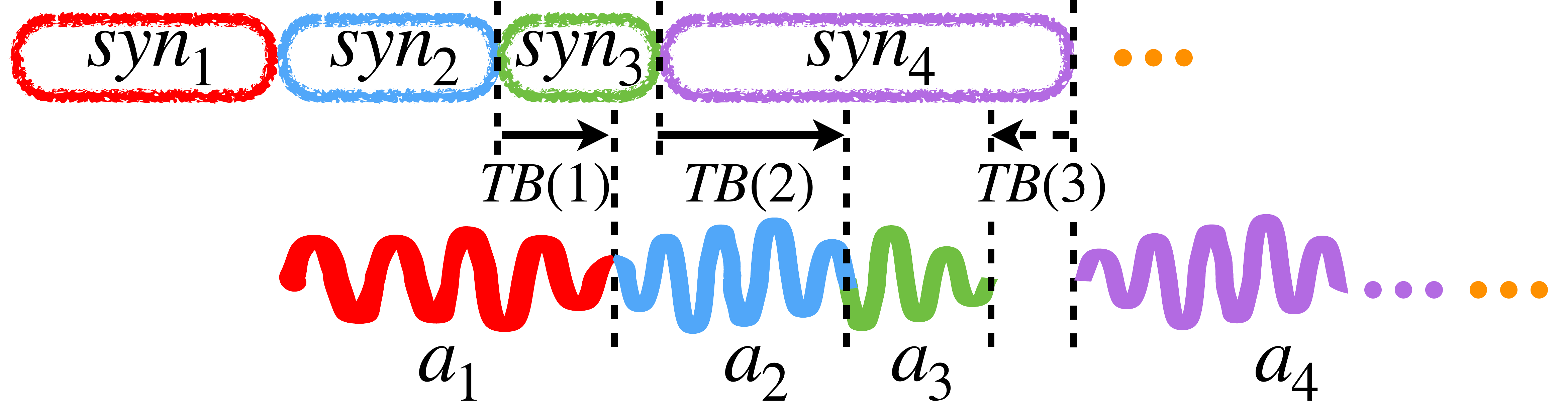}
\caption{
    An example for time balance. The first two steps have positive time balance, implying the first three audio pieces can be played continuously. The third step have negative time balance, meaning that there will be some interruption after the third piece.
}
\label{fig:tb}
\end{figure}

\paragraph{Continuity}
We next show that our method is fast enough so that the generated audios can be played continuously without interruption,
i.e., 
the generation of the next audio chunk will finish {\em before} the audio playing of the current chunk ends
(see Fig.~\ref{fig:tb}).
Let $a_t$ be the playing time of the $t^{th}$ synthesized audio chunk, and $syn_t$ be its synthesis time.
We define the time balance $\tb(t)$ at the $t^{th}$ step as follows (assume $\tb(0)=0$): 
$$
\begin{aligned}
\tb(t) & =  \max\{\tb(t-1), 0\} + (a_t - s_{t+1})
\end{aligned}
$$
Intuitively,
 $\tb(t)$ denotes the ``surplus'' time between the end of audio playing of the $t^{th}$ audio chunk and the end of synthesizing the $(t+1)^{th}$ audio piece.
If $\tb(t) \ge 0$ for all $t$,
then the audio of the whole sentence can be played seamlessly. 
Fig.~\ref{fig:diff} computes the time balance at each step for all sentences in the 300-sentence set for English and Chinese.
We find that the time balance is always positive for both languages and both 
policies.


\begin{figure}[t]
\centering
\includegraphics[width=\linewidth]{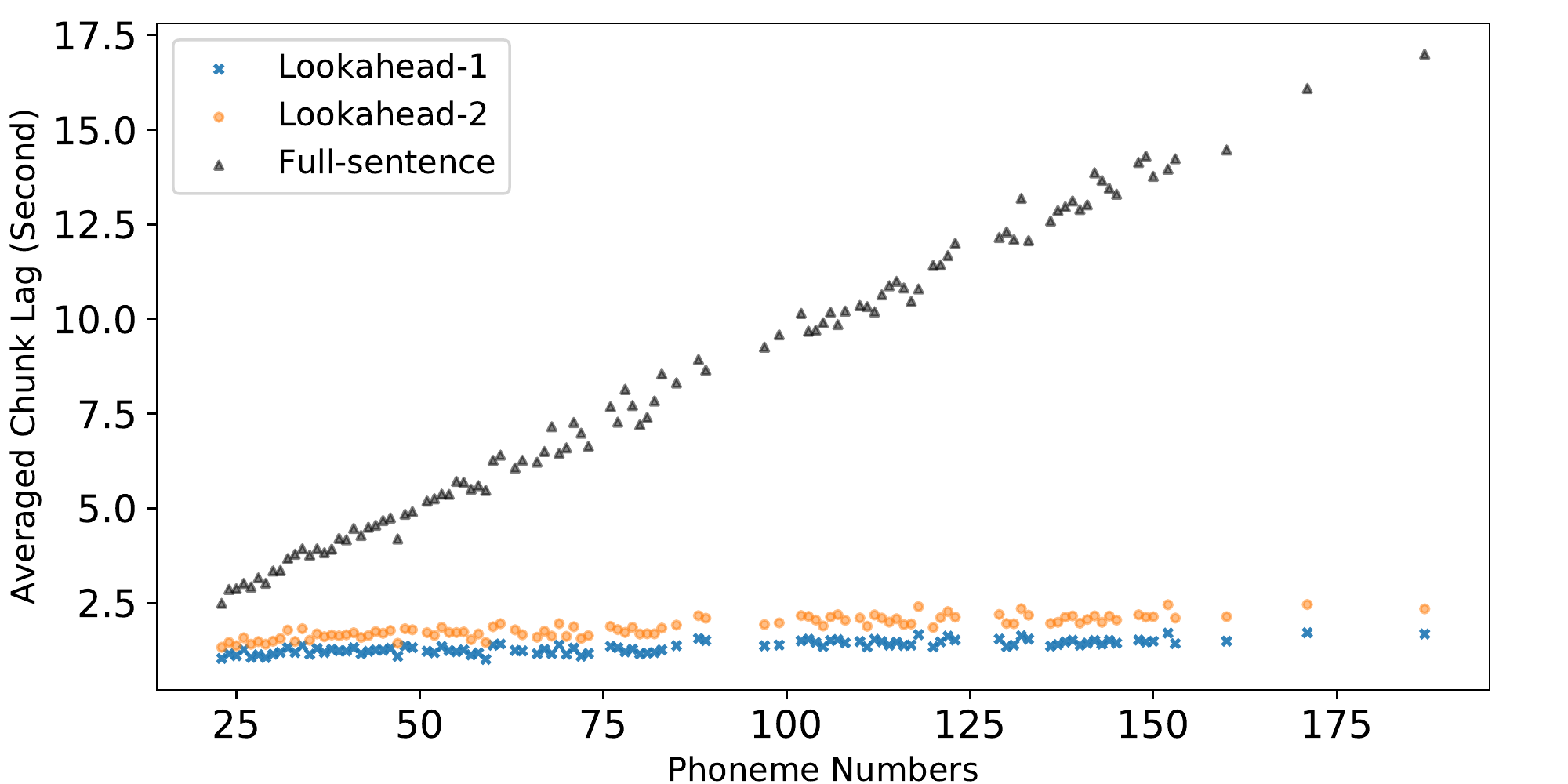}
\includegraphics[width=\linewidth]{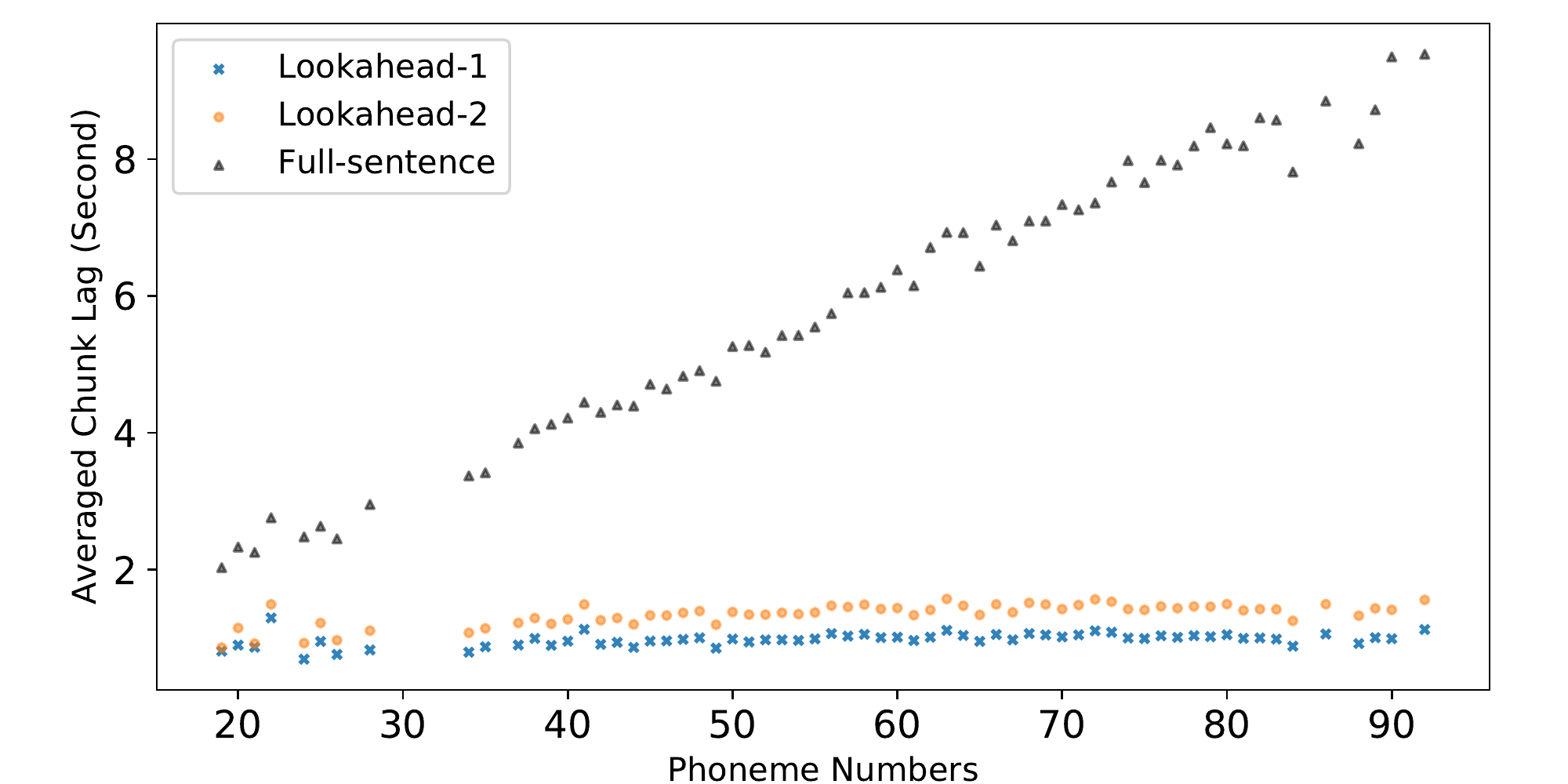}
\caption{Average chunk lag of different methods on English (upper) and Chinese (lower). 
Full-sentence TTS suffers from a delay that increases linearly with the sentence length, while our incremental methods have constant delay. 
}
\label{fig:latency}
\end{figure}


\subsubsection{Input Given Incrementally} 
\begin{figure}
\centering
\includegraphics[width=\linewidth]{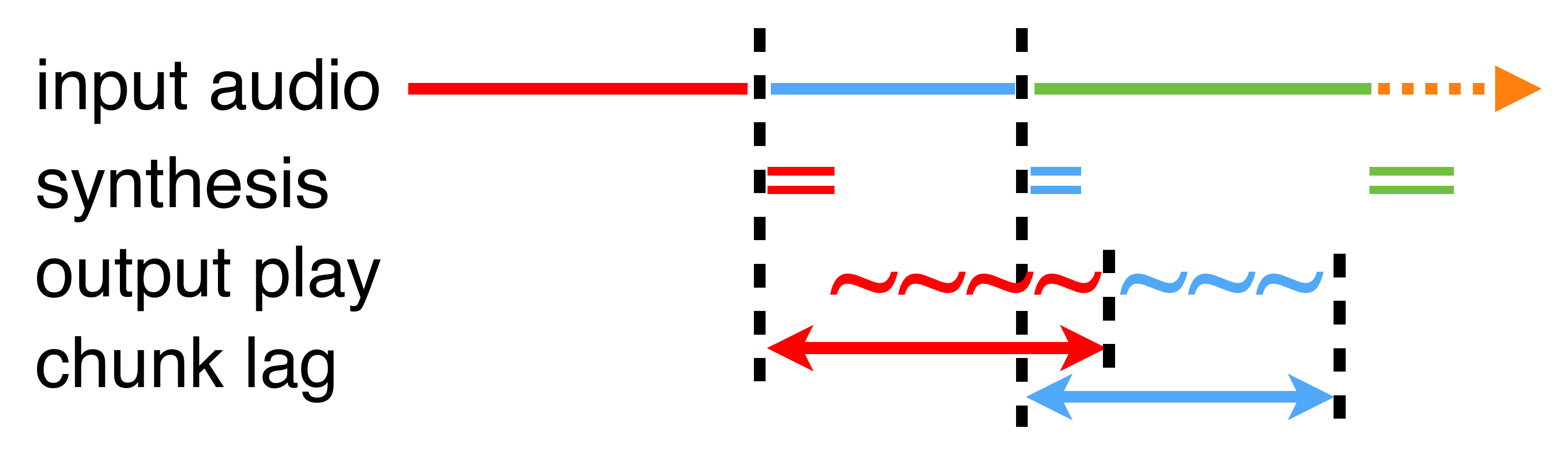}
\caption{
    An example for chunk lags. The arrows represent the lags for different chunks.
}
\label{fig:evs}
\end{figure}

To mimic this scenario, we design a ``shadowing'' experiment
where the goal is to repeat the sentence from the speaker with a latency as low as possible;
this practice is routinely used to train a simultaneous interpreter \cite{lambert:1992}.
For this experiment, our latency needs to include both the computational latency and input latency.
Here we define the {\em averaged chunk lag} as 
the average lag time between the ending time of each input audio chunk
and the ending time of the playing of the corresponding generated audio chunk (see Fig.~\ref{fig:evs}). 

We take the ground-truth audios as inputs and
extract the ending time of each chunk in those audios by the Montreal Forced Aligner~\cite{mcauliffe+:2017}.
The ending time of our chunk can be obtained by combining the generation time, audio playing time and input chunk ending time.
We average the latency results over sentences with the same length and the results are provided in Fig.~\ref{fig:latency}. 

We find that the latency of our methods is almost constant for different sentence lengths, which is under 2.5 seconds for English and Chinese;
while the latency of full-sentence method increases linearly with the sentence length. 
Compared with Fig.~\ref{fig:gens},
larger latency is expected due to input latency.


\section{Conclusions}
We have presented a prefix-to-prefix inference framework for incremental TTS system, and a lookahead-$k$ policy that the audio generation is always $k$ words behind the input.
We show that this policy can achieve good audio quality compared with full-sentence method but with low latency in different scenarios: when all the input are available and when input is given incrementally.

\bibliography{main}
\bibliographystyle{acl_natbib}

\end{document}